# ALPRs - A New Approach for License Plate Recognition using the SIFT Algorithm

F.A. Silva[1,3], A.O. Artero[2], M.S.V. Paiva[3] and R.L. Barbosa[4]

[1]University of Western São Paulo (Unoeste), Dept. of Computer Science,
Presidente Prudente, São Paulo, Brazil
`chico@unoeste.br`
[2]São Paulo State University (Unesp), School of Science and Technology,
Presidente Prudente, São Paulo, Brazil
`almir@fct.unesp.br`
[3]São Carlos Engineering School (USP), Electrical Engineering Department,
São Carlos, São Paulo, Brazil
`mstela@sc.usp.br`
[4]São Paulo State University (Unesp), Sorocaba, São Paulo, Brazil
`ricardo@cartovias.com.br`

## ABSTRACT

*This paper presents a new approach for the automatic license plate recognition, which includes the SIFT algorithm in step to locate the plate in the input image. In this new approach, besides the comparison of the features obtained with the SIFT algorithm, the correspondence between the spatial orientations and the positioning associated with the keypoints is also observed. Afterwards, an algorithm is used for the character recognition of the plates, very fast, which makes it possible its application in real time. The results obtained with the proposed approach presented very good success rates, so much for locating the characters in the input image, as for their recognition.*

## 1. INTRODUCTION

Nowadays, the intense use of video cameras along highways and also in urban areas has contributed a lot to control traffic and, in most applications, the license plate recognition (LPR) of the vehicles is the main objective in traffic analysis, because it consists in a safe way to distinguish clearly imaged vehicles. In the last years many techniques were proposed [30][13][36][22][23][14] [51][1] for the location, segmentation and recognition of the plates in images and videos. There are several applications that use license plate recognition, such as traffic monitoring on highways [45], access control systems in parking lots [49], intelligent transport systems (ITS) [2] etc. License plate character recognition is a critical task, needing to be done with high accuracy. Otherwise, the identification errors will certainly cause very serious implications, for example, sending traffic tickets to the wrong person. However, due the large amount of images that are taken every time, containing a lot of vehicles, the information extraction from these images has become a very complex task. Thus, over the last years, the search for computational solutions, that it is able to automatically recognize the license plates of vehicles, had generated intense research in the area. In toll squares the images are captured in a relatively controlled way, with the cameras always focusing the region containing the vehicle plate. However, some factors still contribute to complicate the task, such as: weather conditions, variations in lighting and shadows, damaged plates, etc. Then the existent techniques continue unable to solve all the cases. However, recently are available some systems that have achieved some positive results, such as the photographic radar systems, which also operate monitoring the traffic and contributing for to identify irregular vehicles.

The LPR process consists of two main steps: 1) locating license plates and 2) identifying characters. In the first step, possible license plate regions are found based on its features. Typically, the used features are derived from the license plate format and the alphanumeric characters. The main plate features include: shape and symmetry [27]; color [25][39][31][46]; texture [39][8][28]; variance of intensity values [30][12][16]. The main character features include: line [53]; the sign of gradient magnitudes; the aspect ratio of characters [18]; the alignment of characters [50]; the distribution of intervals between characters [43].

Other approaches in this area include the work of [20] and [54] that proposed techniques based on combinations of edge statistics and mathematical morphology. In [1] and [23] is presented a method based on morphology with template matching of the characters. In [39] and [56] is proposed the use of fuzzy logic for segmentation. In [10] and [24] is proposed a method using the Hough transform for plate extracting. In [31] is used neural network for color extraction. In [25] is used a genetic algorithm based segmentation to extract the plate region. In [4] are identified signs in real-time video, using a platform DSP (Digital Signal Processing).

The license plate regions obtained in the first step are then examined in the characters identification step. The two main tasks in the characters identification step are character separation and character recognition. The main techniques used in the first task are: morphology [8][16][43]; projection [17][44]; connected components [39]. In the second task, there are a large number of character recognition techniques in the literature, including: artificial neural networks [31][8][26][42]; genetic algorithms [25]; Markov processes [11]; support vector machine [26]. In [21] is proposed a method for extracting characters without prior knowledge of their position and size in the image; more recently, in [51] is proposed a genetic neural network, that it is a combination of genetic algorithm with a back propagation neural network.

The application of the SIFT algorithm in the LPR process is still little explored. One of the few studies in the literature was presented by [55]. In this work the SIFT algorithm was just used in the character recognition step, since the location of the plate region was made using a traditional approach, based on distribution of vertical edges.

In this paper we present a new approach to the LPR process, which uses the algorithm SIFT (Scale Invariant Feature Transform) [34] [35] to locate the license plate. Different from the most of applications that require invariant features to positioning and rotation, and therefore they ignore these two parameters (also obtained with the SIFT algorithm), in this paper these parameters are used to filter the keypoints that have a realistic chance of belong to the plate region. In the proposed approach, this is done by identifying in the image, the regions that have features found in templates corresponding to the digits 0,...,9 existing in the license plate. After identifying a number, a search is started in their neighborhood, until finding the others characters (A,...,Z and 0,...,9), delineating them. Finally, a character recognition algorithm based on the transition among the pixels of the characters [47] is used for the license plate recognition.

The remaining of this paper is organized as follows. Section 2 describes the SIFT algorithm proposed in [35], which aims to identify and to describe the image keypoints; Section 3 presents a new approach using the image description obtained with the SIFT algorithm to identify the correspondences between the templates and the input images; Section 4 presents some experiments with the implementation of the proposed approach. Finally, Section 5 presents some conclusions and suggestions for future work.

## 2. SIFT ALGORITHM

The identification of homologous points in two images is not a very simple task, there are a lot of researches in the area for its automatic execution [5][15]. In fact, the first difficulty is to find the interesting points (keypoints) in the first image, so that they can be searched in the second image.

Because of the excellent qualities of the SIFT algorithm, it has been used in several applications, including the works presented in [7] to construct panoramic image, in [33] to construct mosaic of aerial image sequence, in [9] for the traffic sign recognition, and others.

In this paper we use the SIFT algorithm to find the interesting points in the image and in the templates (containing the characters 0,...,9), and then to locate the plate of the vehicle. The SIFT algorithm is a very efficient method to identify and to describe image keypoints, which is done by performing a mapping with different views of an object or scene, resulting in a vector with 128 values, describing each image keypoint. Additionally, the SIFT algorithm provides the position $x$; $y$, scale $s$ and orientation $\theta$ [35] for all keypoints. The SIFT algorithm consists in the following steps:

Scale-space extrema detection: The keypoints are detected applying a cascade filtering that identifies candidates that are invariant to scale. The scale-space is defined as a function $L(x,y,\sigma)$ in Equation 1, with an input image $I(x,y)$ [29] [32].

$$L(x,y,\sigma) = G(x,y,\sigma) * I(x,y) \tag{1}$$

where * is a convolution in $x$ and $y$ with the Gaussian $G(x,y,\sigma)$ in Equation 2.

$$G(x, y, \sigma) = \frac{1}{2\pi\sigma^2} e^{-(x^2+y^2)/2\sigma^2} \tag{2}$$

To detect stable keypoint locations in space-scale, in [34] is proposed the use of space-scale extrema in the difference-of-Gaussian (DoG) function convolved with the image $I(x,y)$, resulting in $D(x,y,\sigma)$, which can be computed from the difference of two nearby scales separated by a constant multiplicative factor $k = 2^{1/s}$ (where $s+3$ is the number of blurred images for each octave) as in Equation 3.

$$D(x,y,\sigma) = (G(x,y,k\sigma) - G(x,y,\sigma)) * I(x,y) \tag{3}$$

The DoG is an approximation of the scale-normalized Laplacian of Gaussian $\sigma^2\nabla^2 G$ [32]. The maxima and minima of $\sigma^2\nabla^2 G$ produce the most stable image features [38].

Local extrema detection: From $D(x,y,\sigma)$, in [35] is suggested that the local maxima and minima must be detected by comparing each pixel with its eight neighbors in the current image and nine neighbors in the scale above and below (26 neighbors). SIFT guarantees that the keypoints are located at regions and scales of high variations, which make these locations stable for characterizing the image.

Orientation assignment: The scale of the keypoint is used to select the Gaussian smoothed image $L$, with the closest scale, so that all computations are performed in a scale-invariant manner. The gradient magnitude $m(x,y)$ is computed with Equation 4.

$$m(x, y) = \sqrt{\Delta x^2 + \Delta y^2} \tag{4}$$

where $\Delta x = L(x + 1, y) - L(x - 1, y)$ and $\Delta y = L(x, y + 1) - L(x, y - 1)$. The orientation $\theta(x,y)$ is calculated by Equation 5.

$$\theta(x, y) = \arctan(\Delta y / \Delta x) \tag{5}$$

Keypoint description: The next step is to compute a descriptor for the local image region that is distinctive and invariant to additional variations, such as change in illumination or 3D viewpoint. In [35] is suggested that the best approach is to determine the magnitudes and directions of the gradients around the keypoint location. In this approach the Gaussian image on the keypoint scale is used.

## 2.1. Matching between two images

To find the match between two images it is possible to use the keypoints detected with the SIFT algorithm. In [35] is proved that the best match for each keypoint is found by identifying its nearest neighbor, which is defined minimizing the Euclidean distance to the features vectors.

To avoid an exhaustive search, in [35] is suggested the use of a data structure k-d tree [6], that supports a balanced binary search to find the closest neighbor of the features and the heuristic algorithm Best-Bin-First (BBF) is used for the search.

# 3. ALPRs

The algorithm proposed in this paper, called ALPRs (Automatic License Plate Recognition using the SIFT) can be understood by Figure 1, where we can observe that initially the SIFT algorithm is applied in each template, thus the extracted features from keypoints are stored in a database and the templates didn't need be processed at all times. Then, the SIFT algorithm is applied to input image and the extracted features are compared with features of the template keypoint from the database, to achieve the best matching.

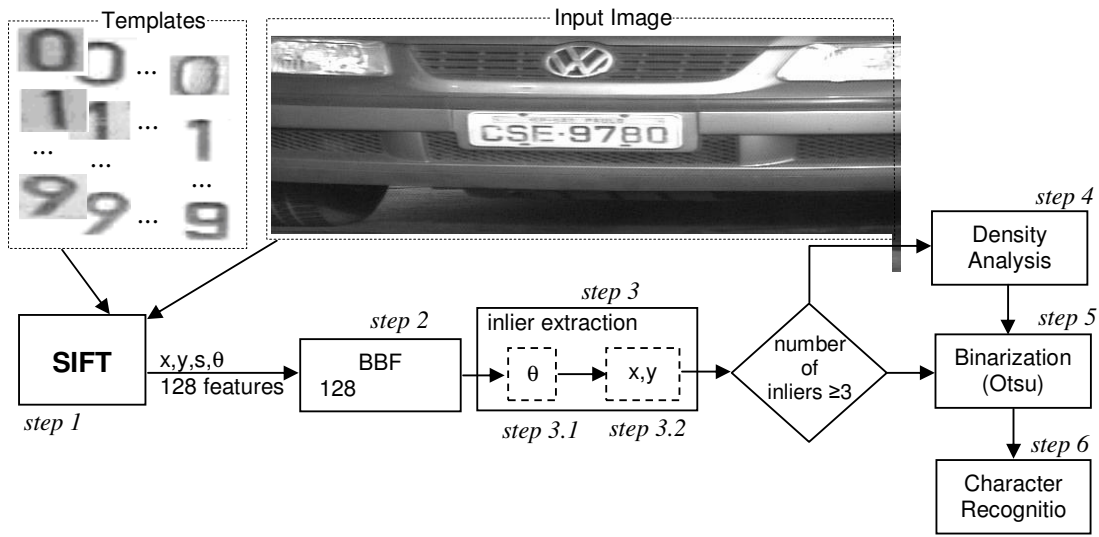

Figure 1. The ALPRs algorithm proposed for the recognition of license plates.

The region around the best matching local is used to determine a thresholding to be used in the image binarization. Finally, after the segmentation of the characters, an algorithm is applied for character recognition [47]. All steps of the algorithm ALPRs are detailed in the following sections.

## 3.1. Plate Identification

License plate identification is done through a search to find the characters on it. In the proposed approach, the SIFT algorithm is applied on the input image and in the set of template images, containing the similar characters to them on the plate. In order that this process doesn't become very slow, only the digits 0,...,9 are searched on the plates. Figure 2 shows the template containing the digits used by the SIFT in this work.

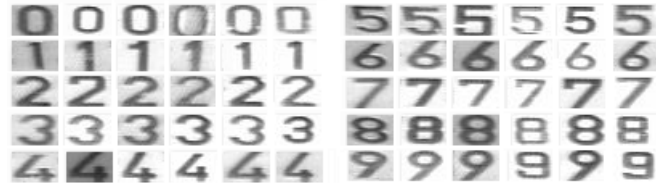

Figure 2. Template images of the numeric digits used for the matching with the input images.

In step 1 of the ALPRs algorithm, the SIFT algorithm is applied on the template representing the digit '3' in Figure 2 (a) and on the input image in (b). The keypoints found, by the SIFT algorithm, in the template and the input image, are shown in Figure 3.

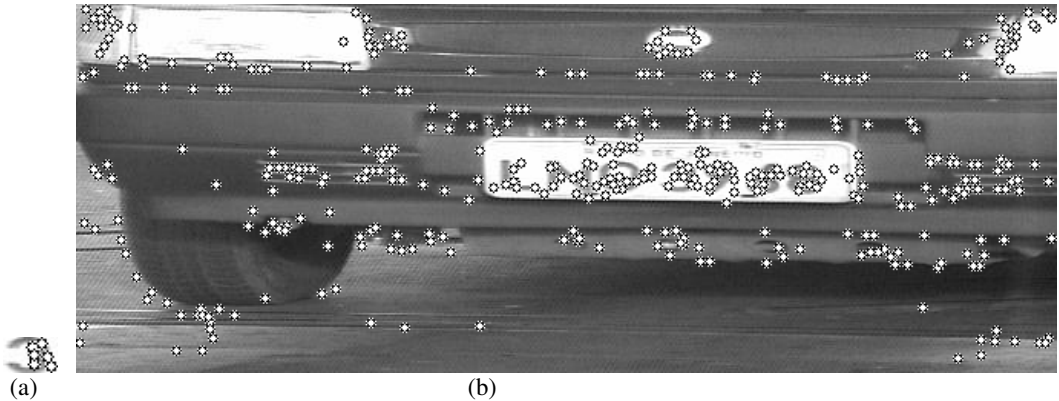

(a)                                                        (b)

Figure 3. Keypoints extracted by the SIFT algorithm in: a) template and b) input image.

In this figure we observe that the SIFT algorithm found several interesting points (keypoints) in the input image in (b), however, nor all of them have similar features to the located points in the template in (a). Thus, in step 2, the ALPRs algorithm eliminates the points that do not have similar features to the located points in the template. This is done by comparing (matching) the features of each template keypoint (Figure 3 (a)) with features of the input image keypoints (Figure 3 (b)) and, at the end, only the keypoints illustrated in the Figure 4 (b) are maintained in the input image.

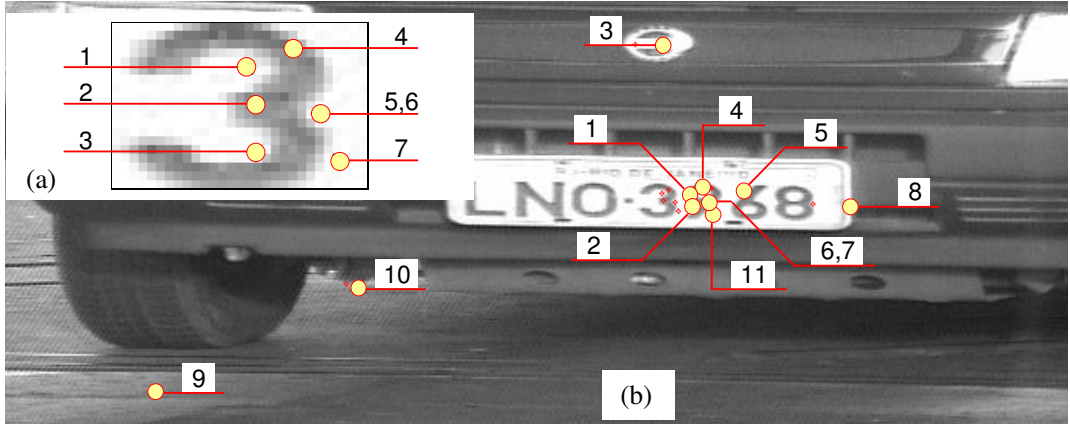

Figure 4. a) Set of template keypoints, b) Set of input image keypoints.

Although most of the keypoints that do not match with the template keypoints have been eliminated from the input image at this step, there are still some points that visually do not correspond with the template, such as the points: 3, 5, 8, 9 and 10. Thus, the algorithm ALPRs in your step 3, tries to identify only the inliers (data points that fit a particular model within a error tolerance), which is done in two stages: 1) comparing the orientation angles of the local gradients, 2) calculating the correlation among the keypoints spatial coordinates in the input image and in the template.

Table 1 presents the values obtained from SIFT algorithm for the parameters of spatial coordinates ($x,y$) and orientation θ, for the template keypoints (Figure 4 (a)), and of the input image keypoints (Figure 4 (b)).

Table 1. Keypoints from Figure 4, with the coordinates ($x,y$) and orientation θ, after computing the matching SIFT features.

| Template keypoints | | | | Input image keypoints | | | |
|---|---|---|---|---|---|---|---|
| # | x | y | θ | # | x | y | θ |
| 1 | 15 | 5 | -1,6244128 | 1 | 395 | 108 | -1,7278086 |
| 2 | 16 | 9 | 2,3460571 | 2 | 396 | 112 | 2,3459965 |
| 3 | 16 | 14 | -2,3614827 | 3 | 379 | 24 | 1,2154149 |
| 4 | 20 | 3 | 1,0139291 | 4 | 399 | 106 | 1,1906057 |
| | 20 | 3 | 1,0139291 | 5 | 425 | 105 | 1,3271831 |
| 5 | 23 | 10 | 0,6453162 | 6 | 403 | 113 | 0,6453041 |
| 6 | 23 | 10 | -0,5482631 | 7 | 403 | 113 | -0,5482711 |
| | 23 | 10 | -0,5482631 | 8 | 486 | 114 | -0,4185706 |
| 7 | 25 | 15 | -0,3427376 | 9 | 89 | 217 | -1,7855562 |
| | 25 | 15 | -0,3427376 | 10 | 205 | 159 | -1,7662604 |
| | 25 | 15 | -0,3427376 | 11 | 405 | 118 | -0,3422799 |

In this table we observed that a template keypoint can match more than one input image keypoint, as it happens with the keypoints: 4 from template that has similar features to the keypoints features 4 and 5 from input image; 6 from template with 7 and 8 from input image; 7 from template with points 9, 10 and 11 from input image.

The SIFT algorithm proposes features that are invariant regarding to scale, rotation and differences of 3D viewpoint. However, in this work, we don't hope to find rotations above 10 degrees between the template and the input image. In this case, it is proposed that the process of inliers extraction compares the directions of local gradients θ (also obtained with the SIFT algorithm – Equation 5 [35]). Given n keypoints $K_i^j$ ($j=1,..,n$) in the input image and its corresponding $K_i$ in the template, the pair $K_i$ and $K_i^j$ are considered candidates to inliers if the $K_i\theta$ and $K_i^j\theta$, the orientation of $K_i$ and $K_i^j$, respectively, satisfies $\left| K_i\theta - K_i^j\theta \right| \leq 2\pi/36$. Otherwise, they are immediately classified as outliers (outliers: data points that do not fit a particular model within an error tolerance) and removed form the set of inliers candidates.

Following this strategy with the orientation values in Table 1, it is verified that the orientations are similar only for the pairs of points (template-image): 1-1; 2-2; 5-6; 6-7; 6-8; 7-11. Thus, the pairs of template-image, which do not have similar orientation are considered outliers, and they are discarded in the next steps. After that, the set of input image keypoints is reduced to six points {1, 2, 6, 7, 8 and 11}, shown in Figure 5.

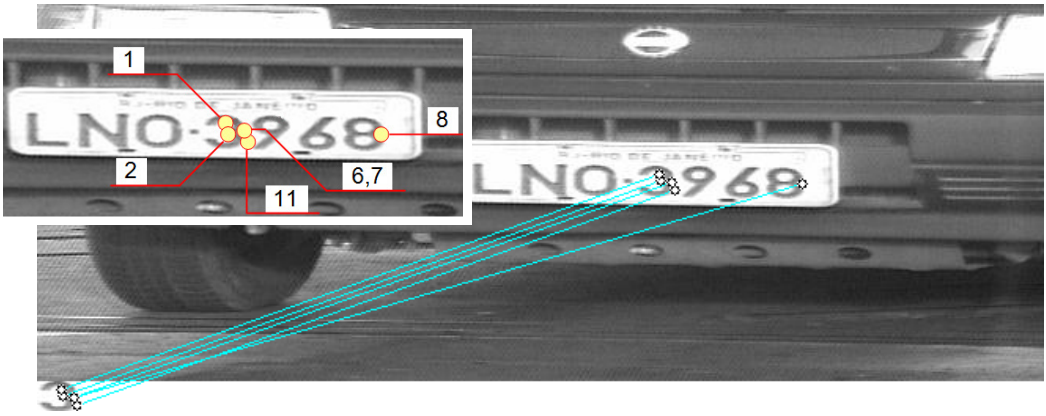

Figure 5. Keypoints candidates to inliers after elimination of the outliers.

In step 3.2, the ALPRs algorithm searches the region with the highest keypoints concentration in the input image, taking into account the spatial coordinates of the template keypoints. This search is performed to find the region with the highest density of candidates for the inliers. The density function of a data matrix $D_{mxn}$ for an n-dimensional variable $x$, based on a kernel density estimation function $K$, is defined in [48] with Equation 6.

$$f(x) = \frac{1}{mh^n} \sum_{i=1}^{m} K\left( \frac{x - d_i}{h} \right) \qquad (6)$$

where $d_i$ is the $i$-th record of the data set (the n-dimensional variable given by the $i$-th row in matrix $D$) and $K$ is the kernel function defined for the n-dimensional variable $x$, which satisfies $\int_{R^n} K(x)dx = 1$. Even though various kernels have been proposed, the most common ones are square wave and Gaussian functions. The parameter $h$ in the density function defines a smoothing factor or bandwidth. When $h\rightarrow 0$ one obtains a sum of Dirac's Delta functions [48], which tend to highlight only the overlapping of data records, while large $h$ values highlight areas with large concentration of data, i.e., clusters. An alternative computation, more efficient than using Equation 6, considers only the influence of neighboring points on each point in the data set, for an arbitrary neighborhood. This may be modelled by a mathematical function referred to as influence function, which describes the impact of a point on its neighborhood and is equivalent to a smoothing filter. The point density is approximated by the sum of the influence functions of all its neighboring elements within a given region [19]. As in [3] we employed this approach in our solution, using a square wave filter function rather than a Gaussian. After the determining the region of highest density of points candidates, the inliers points are finally determined, and the keypoint 8 (Figure 5) was eliminated from the set of input image keypoints (reduced to five points {1, 2, 6, 7 and 11}).

It is important to observe that, to be characterized a reasonable matching between one of the templates and an input image, it is necessary and sufficient that it is found a minimum amount of inliers, that empirically has been defined as a minimum of three inliers. In the cases that are not found at least three inliers for any of the templates used in the search, an additional strategy is used (step 4) that consists in analyzing, on the input image, all inliers obtained individually using all the templates, and locating the region of higher density. Figure 6 shows a case in that the input image did not reach the minimum of three inliers with a same template (each template had up two inliers). In this figure are presented all candidates to inliers that have been approved in step 3.1, obtained using all the templates. In this case, the region that should contain some of the plate characters is one that presents the highest density to inlier candidates. In Figure 6, this region is around the keypoints on the digit '3'.

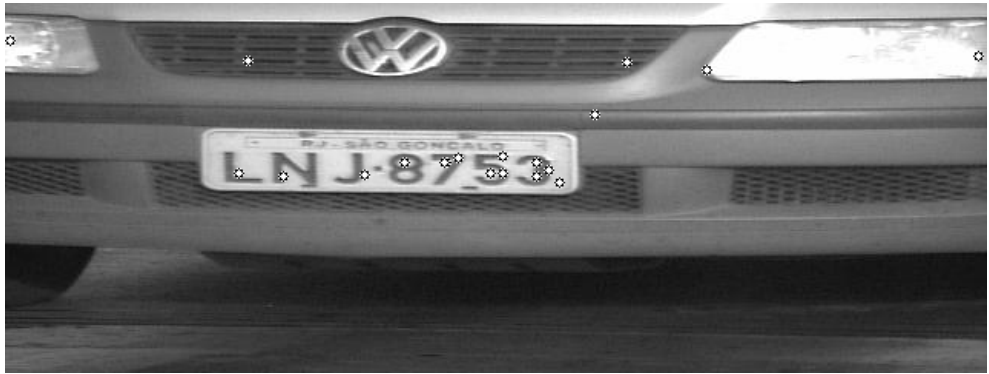

Figure 6. Set of inliers obtained with all the templates.

At this time, the region of the plate is identified and, subsequently, it is necessary to segment the characters, so that it is possible to perform their recognition. In this paper it is done in step 5, using a segmentation method based in the Otsu thresholding [41]. The threshold value must be obtained just considering the region of the plate. The Otsu method is based in a discriminating analysis, where the threshold value is obtained assuming that the image pixels can be classified into two classes ($C_0$ e $C_1$) that are the object and the background. Taking $\sigma_B^2$ and $\sigma_T^2$ the variances between the classes and total, respectively. The optimal value for the threshold, according to this method, is obtained by minimizing n in Equation 7.

$$n = \frac{\sigma_B^2}{\sigma_T^2}$$
(7)

## 3.2. Alphanumeric characters Recognition

The character recognition (letter or number) is an important area of pattern recognition [40] with several applications in automation and information treatment [52]. In this paper, only the numeric digits (0,1,..,9) are used in the plate identification (Section 3.1), however for the plate recognition, all the characters are used (0,1,..,9 and A,B,…,Z). The solution adopted in the step 6 of the ALPRs algorithm (Figure 1) was developed in [47], because it provides a high performance and its recognition time is adequate for real time tasks. This algorithm samples the characters images in a grid and it observes the transitions between the pixels values (0 and 1 – binary images) of adjacent pixels. In this way, a character image sampled in a grid with dimensions *m* x *n* generates a transition vector with *m.n* values. The transitions are defined on the image character using the path illustrated in Figure 7.

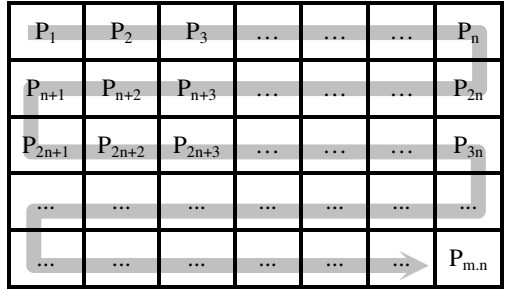

Figure 7. Pixels sequence in the image used to model the behaviour of the character.

After investigating the behavior of these transitions to a set of characters from a given class (supervised training), is defined a set of rules that will be used later, to classify the characters whose classes are not known. In [47] is suggested the use of a list structure to store the transitions that don't occur between each adjacent pair of pixels in the image of each character in their respective classes. As the images used are binary, the possible transitions between two adjacent pixels are: $\underline{00}$, $\underline{01}$, $\underline{10}$ e $\underline{11}$. Figure 8 (a) shows the images of two 'A' characters (class 1) and two 'C' characters (class 2). In (b) it has a graph, using parallel coordinates, which illustrates the transitions between adjacent pixels of the images in (a). In (c) it has a list with the transitions that do not occur between adjacent pixels in these images, where it is possible to identify that between the attributes $a_3$ and $a_4$ occur the transitions $\underline{01}$ and $\underline{11}$, and the transitions $\underline{10}$ and $\underline{00}$ do not occur. While the class 2 just possesses the transition $\underline{10}$ between the same attributes.

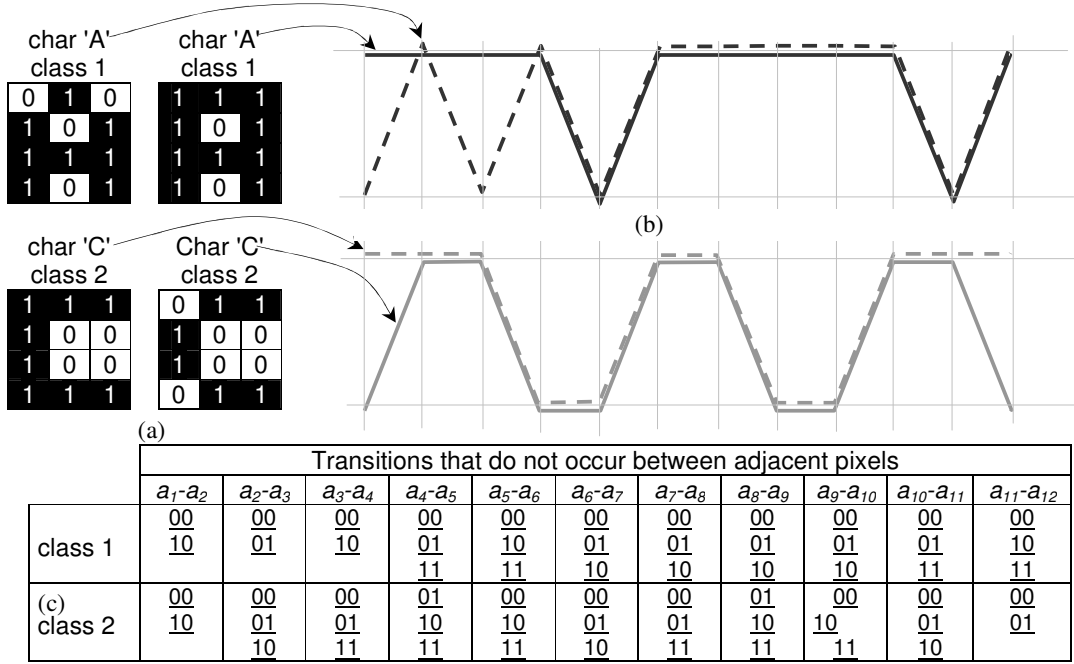

Figure 8. a) Images representing two 'A' characters in the class 1 and two 'C' character in the class 2; b) Exhibition of the records of the class 1 in black and of the class 2 in gray in parallel coordinates; c) Transitions do not occur among adjacent attributes.

The list structure in Figure 8 (c) has 30 restrictions for class 1 and 31 restrictions for class 2. After obtaining all the transitions that do not occur between adjacent attributes in each class, the characters can be classified in the class that presents the lowest number of inconsistencies in relation to characteristic transitions annotated in each class. The three rules for the character classification are:

- $R_1$ – Recognize the character in the class whose transitions are not violated by the transitions in the class;

- $R_2$ – Having more than one class that satisfies this condition, the class that has the most restrictions (more restrictive class) it should be the chosen;

- $R_3$ – When transitions do not satisfy all the restrictions of any class, the character may be inserted in the class less violated, or to be classified as noise in extreme cases.

The last rule ($R_3$) requires a user-defined threshold to determine when the character should be classified as noise. For example, when it does not meet at least 30% of the restrictions of any class rules.

Two strategies are also presented in [47], to increase the quality of the classifications: 1) Accomplishing a denser sampling of the characters, that it will increase the number of transitions between adjacent pixels; 2) Using the transitions among three pixels instead of transitions between two adjacent pixels. Thus, the eight transitions to be checked are: 000, 001, 010, 011, 100, 101, 110 e 111.

## 4. EXPERIMENTS

This section presents some results of license plates character identification and recognition, using the ALPRs algorithm proposed in this paper, comparing with two other approaches presented in the literature [1][37]. It is necessary to observe that these techniques for the identification and recognition of the characters use a step of training of the templates, thus, a comparison using untrained images becomes inadequate. To avoid this problem, in the two approaches used in this comparison we used the original images from their papers. Both experiments were performed in a computer with an Intel Core I3 (2.13 GHz) processor.

### 4.1. Experiment 1

In order to train the recognition algorithm were used the set of character presented in Figure 9, which are very similar to those adopted in the license plates in Brazil.

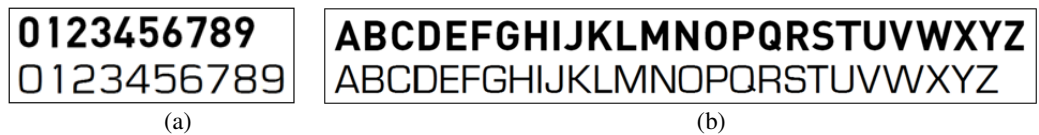
(a)                                        (b)
Figure 9.  Set of characters used in classifier training of the Experiment 1.

In order to apply the recognition algorithm based on the transitions between adjacent pixels, each character was sampled into a grid with 65 x 60 pixels. In this way, each character was represented by 3,900 transitions.

In this experiment, we used 60 frontal images (640 x 240 pixels) of vehicles, obtained in toll squares.  These images are available by the Laboratory of Digital Signal Processing and Images of the Brazilian Center of Physical Researches, and can be freely downloaded from www.cbpf.br/cat/pdsi/lpr/lpr.html. Figure 10 shows some examples in which the plates are correctly recognized. The first column shows some input images used in the experiment and the second and third columns show the results of the license plate recognition.

| | | |
|---|---|---|
| 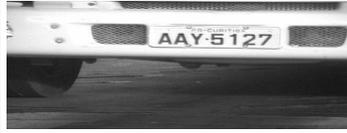 | 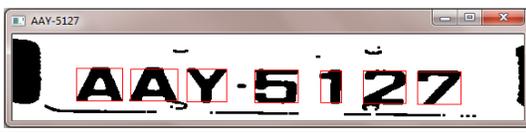 | Identified plate: AAY-5127 |
| 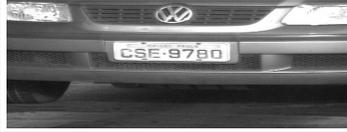 | 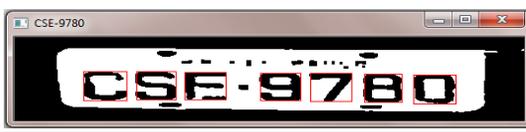 | Identified plate: CSE-9780 |
| 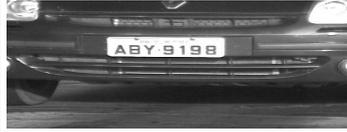 | 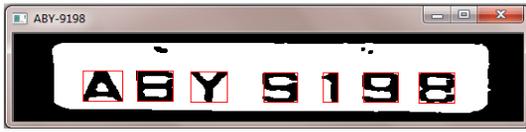 | Identified plate: ABY-9198 |
| 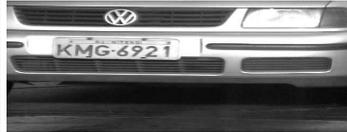 | 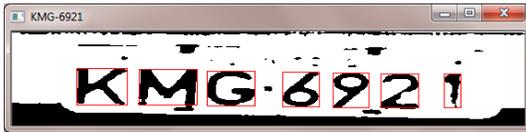 | Identified plate: KMG-6921 |
| 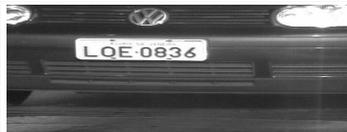 | 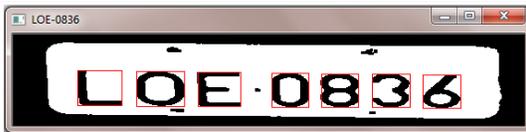 | Identified plate: LOE-0836 |
| 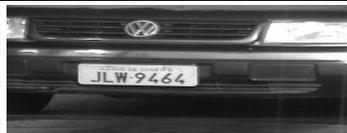 | 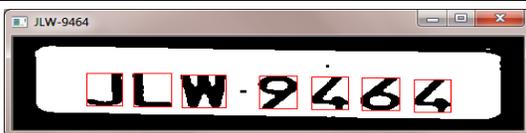 | Identified plate: JLW-9464 |
| 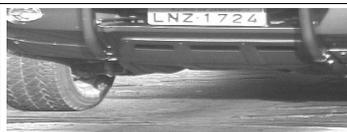 | 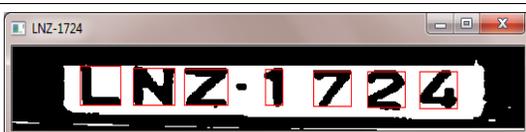 | Identified plate: LNZ-1724 |

Figure 10. Satisfactory results obtained by applying the proposed approach.

In spite of the screw holes for fixing the plates were segmented by the Otsu thresholding in the same class of the characters, the ALPRs algorithm was able to delineate the region containing some of the characters, and then, through a left to right scan, it was able to locate the remaining characters.

However, seven of the 60 plates, with a poor image quality, were not recognized, as illustrated in Figure 11. The results (a) and (b) presented a failure in the segmentation step. In (a), the character 'L' was poorly segmented and it can not be considered a valid character. In (b) the same happened with the character '6'. The character '9' in (b) and all characters in (c) to (g), were properly segmented, however, the recognition algorithm was not successful.

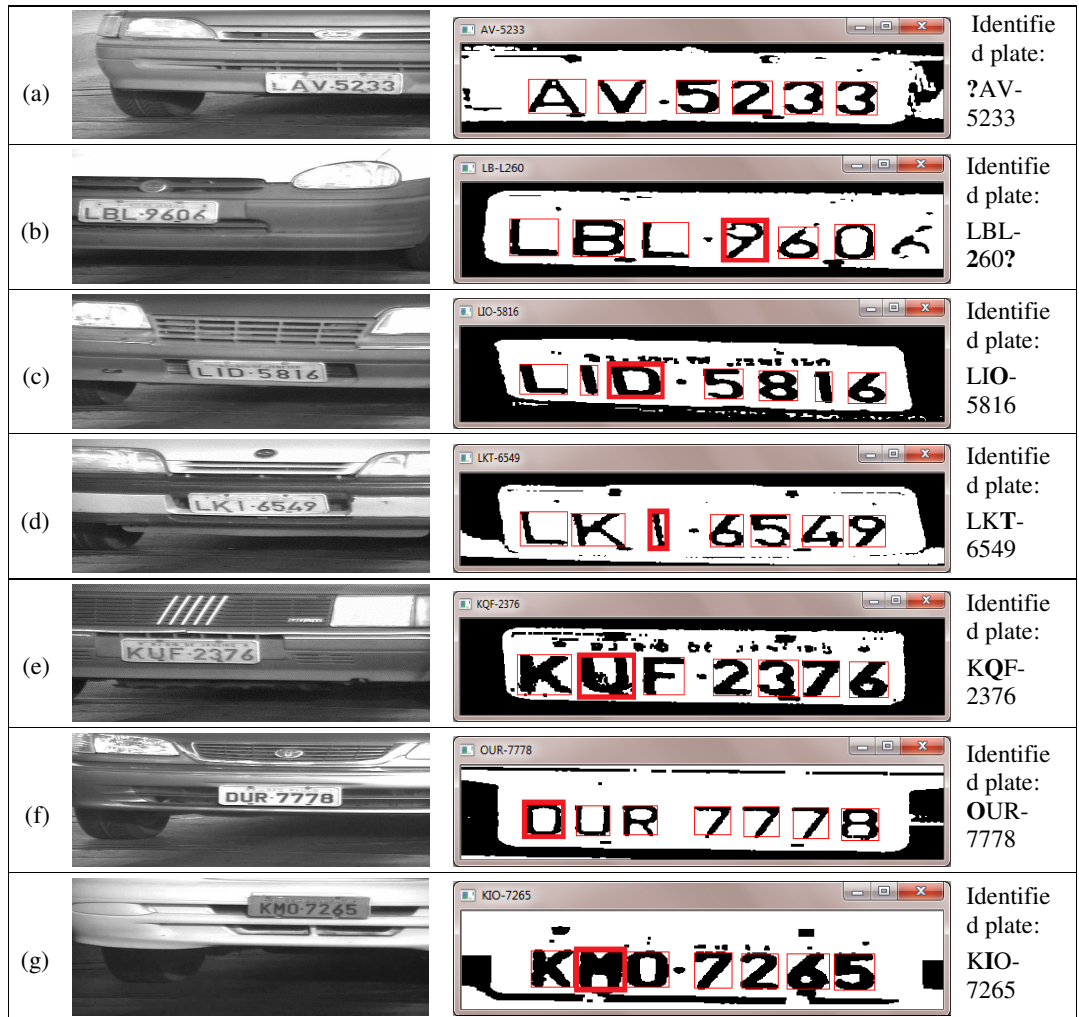

Figure 11. Results showing all errors obtained with the ALPRs algorithm.

In this experiment, the success rate obtained by [1] was 78%, while using our approach was 88.33% (53 successes in a total of 60 plates). It should be observed that the hit rate of the locating characters process was 99.52%, because 418 of the 420 existent characters in the 60 plates were properly located. On the other hand, the hit rate of the character recognition step was 98.57% (414 successes in a total of 420 characters). The average time for the plate recognition was 5.1716 seconds. The processing time for each plate, in each stage of the ALPRs algorithm are: 5.0115 seconds for the execution of the SIFT and the matching algorithm; 0.0907 seconds for segmentation (threshold, clipping of the characters); 0.0694 seconds for the characters recognition using their transitions.

## 4.2. Experiment 2

In this second experiment, the set of character used to train the recognition algorithm is presented in Figure 12 (sampled into a grid with 50 x 30 pixels - 1,500 transitions).

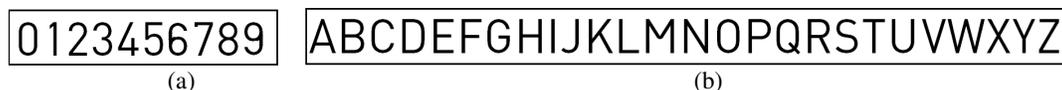

(a)                    (b)

Figure 12. Set of characters used in classifier training of the Experiment 2.

The set of vehicle image used in this experiment, from [37], are available at javaanpr.sourceforge.net. We used 60 images with different sizes from this set to accomplish the experiments. Figure 13 shows some satisfactory results by applying the proposed approach. In this experiment, the success rate obtained by [37] was 73.33%, while our approach had a hit rate of 70% (42 successes in a total of 60 plates). The hit rate of the locating characters process was 93.33% and the hit hate of the character recognition step was 88.10% (370 successes of 420 characters).

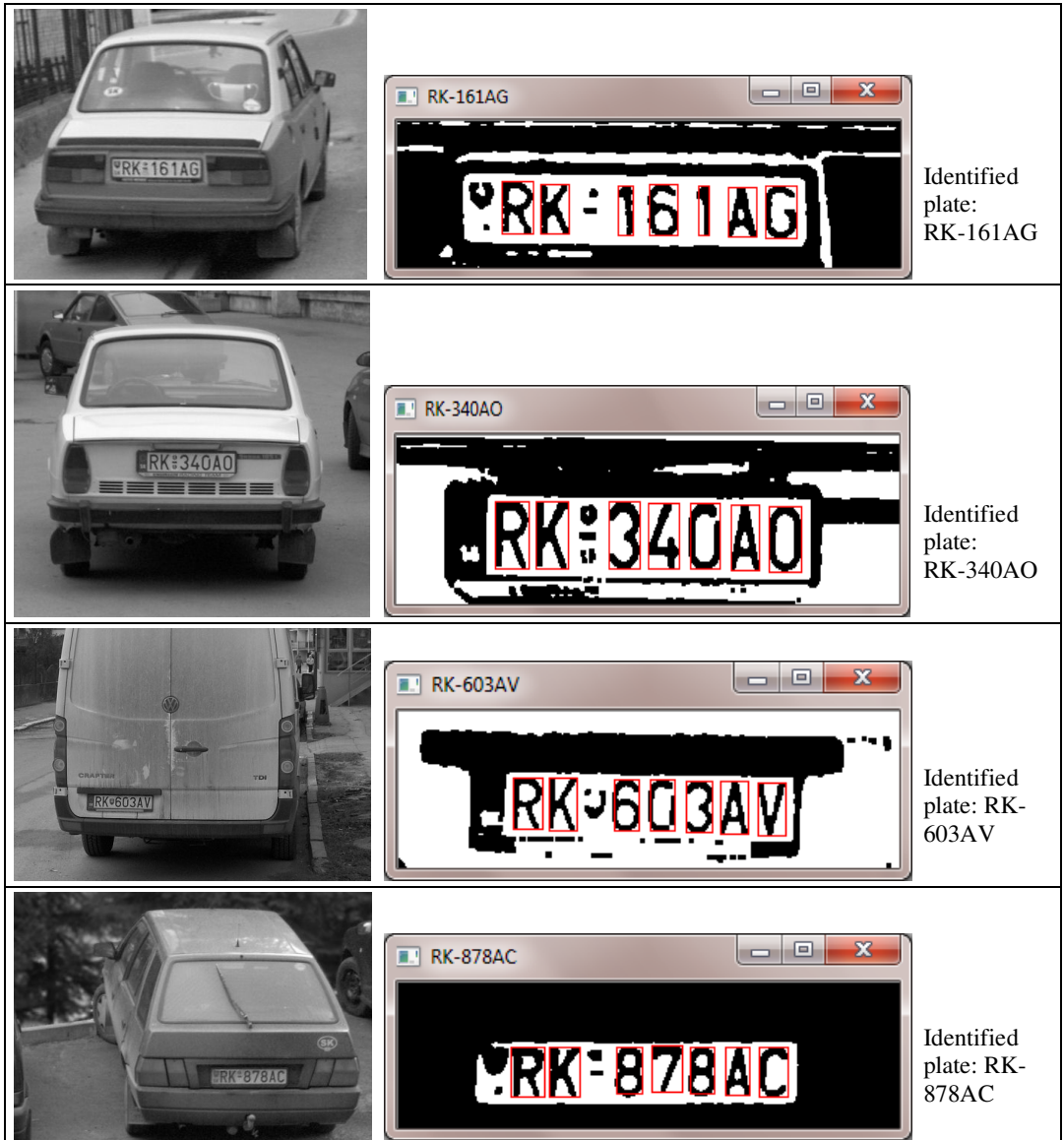

Figure 13. Same good results with the proposed approach.

The average time for the plate recognition was 6.7918 seconds (6.6594 seconds for the execution of the SIFT and the matching algorithm, 0.1116 seconds for segmentation, 0.0208 seconds for the characters recognition).

# 5. CONCLUSIONS AND FUTURE WORK

This paper presented a new algorithm for license plate recognition, which uses the SIFT algorithm to locate the plate in the images. The SIFT has been shown a great algorithm for finding keypoints features in the images and that it has been able to obtain satisfactory results in the LPR process. In this work, these keypoints were used in the matching process between input images and characters templates.

A particular advantage of the approach proposed in this paper is the possibility to change the techniques used in the steps of the algorithm ALPRs, in order to improve the results more and more, by increasing the success rates, and decreasing the processing times. An analysis of the processing time in the Experiment 1 (Section 4.1), showed that the SIFT and the matching algorithms spend 96.90% of the total processing time. In the Experiment 2 (Section 4.2), the processing time for these tasks was 98.05%. Thus, for real time applications, we need to investigate other solutions for this task. Other characters recognition algorithm should be investigated to improve the hit rate of the character recognition step.